\documentclass[11pt]{article}
\usepackage{coling2020}
\usepackage{times}
\usepackage{url}
\usepackage{latexsym}
\usepackage{graphicx}
\usepackage{color}

\usepackage{bm}
\usepackage{algorithm}
\usepackage{algpseudocode}
\usepackage{amsmath}

\colingfinalcopy 

\title{Multilingual Neural RST Discourse Parsing}

\author{Zhengyuan Liu, \ Ke Shi, \ Nancy F. Chen \\
  Institute for Infocomm Research, A*STAR, Singapore \\
  \texttt{\{liu\_zhengyuan,shi\_ke,nfychen\}@i2r.a-star.edu.sg}}

\date{}

\begin{document}
\maketitle

\begin{abstract}
Text discourse parsing plays an important role in understanding information flow and argumentative structure in natural language. Previous research under the Rhetorical Structure Theory (RST) has mostly focused on inducing and evaluating models from the English treebank. However, the parsing tasks for other languages such as German, Dutch, and Portuguese are still challenging due to the shortage of annotated data.
In this work, we investigate two approaches to establish a neural, cross-lingual discourse parser via: (1) utilizing multilingual vector representations; and (2) adopting segment-level translation of the source content. Experiment results show that both methods are effective even with limited training data, and achieve state-of-the-art performance on cross-lingual, document-level discourse parsing on all sub-tasks.
\newline
\end{abstract}

\section{Introduction}
\label{introduction}
\begin{figure}[t]
    \begin{center}
      \includegraphics[width=16cm]{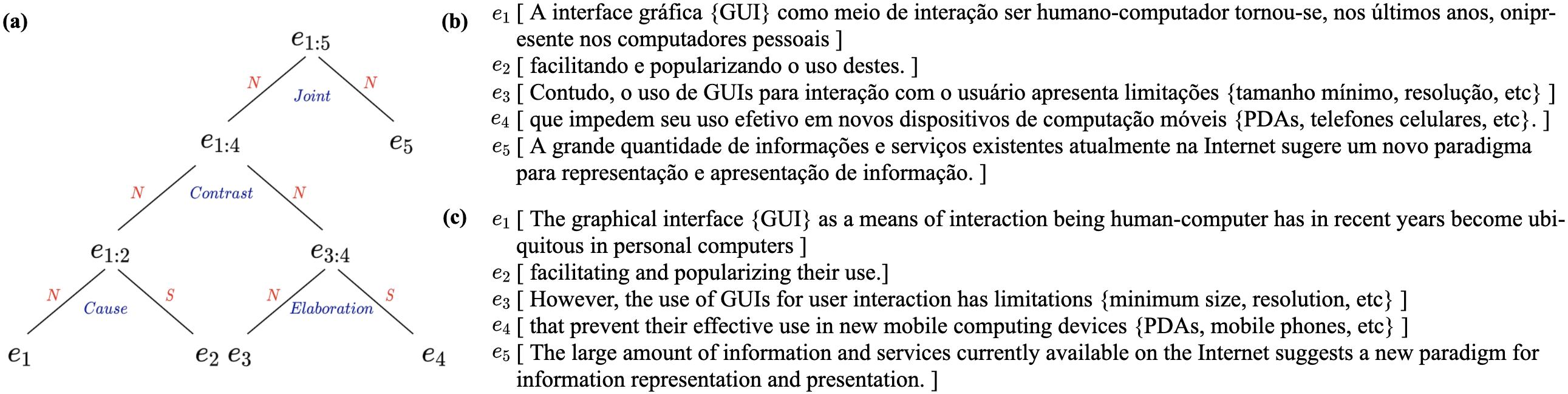}
    \end{center}
    \caption{(a) An RST discourse tree example. $e_i$, $e_{j:k}$, $N$ and $S$ denote elementary discourse units, spans, nucleus and satellite respectively; (b) The original EDU segments in Portuguese; (c) The translated EDU segments in English.}
    \label{fig:RST_TREE}
\end{figure}

\blfootnote{
    %
    %
    %
    %
    %
    %
    \hspace{-0.65cm}  
    This work is licensed under a Creative Commons 
    Attribution 4.0 International License.
    License details:
    \url{http://creativecommons.org/licenses/by/4.0/}.
}
Rhetorical Structure Theory (RST) \cite{mann1988rhetorical} is one of the most influential theories of discourse analysis, under which a document is represented by a hierarchical discourse tree. As shown in Figure \ref{fig:RST_TREE}a, the leaf nodes of an RST tree are text spans named Elementary Discourse Units (EDUs), and the EDUs are connected by rhetorical relations (\emph{e.g.,} \textit{Cause}, \textit{Contrast}) to form larger text spans until the entire document is included. 
The rhetorical relations are further categorized to \textit{Nucleus} (core part) and \textit{Satellite} (subordinate part) based on their relative importance. Thus, document-level discourse parsing consists of three sub-tasks: tree construction, nuclearity determination and relation classification. Moreover, downstream natural language processing tasks can benefit from RST-based structure-aware document analysis, such as summarization \cite{liu-chen-2019-exploiting,xu-etal-2020-discoBert} and machine comprehension \cite{gao2020discoReading}.

By utilizing various linguistic characteristics (e.g., $N$-gram and lexical features, syntactic and organizational features), statistical approaches have obtained substantial improvement on the English RST-DT benchmark \cite{sagae2009analysis,hernault2010hilda,joty2013combining,li2014text,heilman2015fast}. Recently, neural networks have been making inroads into discourse analysis frameworks, such as attention-based hierarchical encoding \cite{li2016discourse} and integrating neural-based syntactic features into a transition-based parser \cite{yu2018transition}. Lin et al. \shortcite{lin2019unified} and their follow-up work \cite{liu2019hierarchical} successfully explored encoder-decoder neural architectures on sentence-level discourse analysis, with a top-down parsing procedure.

Although discourse parsing has received much research attention and progress, the models are mainly optimized and evaluated in English. The main challenge is the shortage of annotated data, since manual annotation under the RST framework is labor-intensive and requires specialized linguistic knowledge. For instance, the most popular benchmark English RST-DT corpus \cite{carlson2002rst} only contains 385 samples, which is much smaller than those of other natural language processing tasks. The treebank size of other languages such as German \cite{stede2014-MAZ-corpus}, Dutch \cite{redeker2012-dutch-rst-dt} and Basque \cite{iruskieta2013-basque-rst-dt} are even more limited. Such limitations make it difficult to achieve acceptable performance on these languages required to fully support downstream tasks, and also lead to poor generalization ability of the computational approaches.

Since the treebanks of different languages share the same underlying linguistic theory, data-driven approaches can benefit from joint learning on multilingual RST resources \cite{braud-etal-2017-cross-lingual}. Therefore, in this paper, we investigate two methods to build a cross-lingual neural discourse parser: (1) From the embedding perspective: with the cross-lingual contextualized language models, we can train a parser on the shared semantic space from multilingual sources without employing a language indicator;
(2) From the text perspective: since each EDU is a semantically-cohesive unit, we can unify the target language space by EDU-level translation, while preserving the original EDU segmentation and the discourse tree structures (see Figure \ref{fig:RST_TREE}c). To this end, we adapted and enhanced an end-to-end neural discourse parser, and investigated the two proposed approaches on 6 different languages. While the RST data for training is still in a small scale, we achieved the state-of-the-art performance on all fronts, significantly surpassing previous models, and even approaching the upper bound of human performance.
Moreover, we conducted a topic modeling analysis on the collected multilingual treebanks to evaluate the model generality across various domains.

\section{Methodology}
\subsection{Neural Discourse Parser}
\label{ssec:neural_parser}
Since the encoder-decoder neural architecture with a top-down parsing procedure proposed in \cite{lin2019unified} has achieved impressive performance on sentence-level discourse analysis, here we adapted and enhanced it on the document-level parsing task. The neural model consists of an encoder, a span splitting decoder, and a nuclearity-relation classifier.

\noindent\textbf{Encoder:}
The encoder produces EDU-level representations via a hierarchical encoding process.
Given a document containing $n$ tokens, an embedding layer is used to obtain token-level representations $\widetilde{T}=\{\widetilde{t}_1,...,\widetilde{t}_n\}$. Then we obtain EDU-level representations by averaging the token embeddings for each EDU, and feed them to a Bi-GRU \cite{cho-etal-2014-BiGRU} component for document-level context-aware modeling. Moreover, to exploit implicit syntactic information like part-of-speech (POS) and sentential information \cite{yu2018transition}, we incorporate boundary embeddings at both ends of each EDU from $\widetilde{T}$ to the context-aware vectors, and obtain the final EDU representation $E=\{e_1,...,e_m\}$, where $m$ is the total EDU number.

\begin{figure}
    \centering
    \includegraphics[width=15cm]{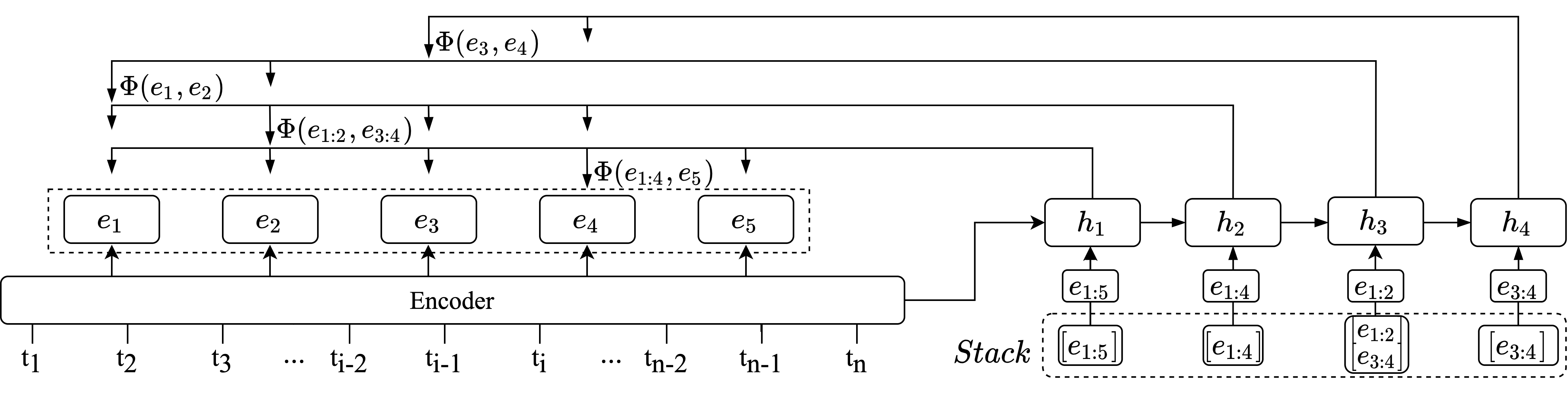}
    \caption{Document-level neural parser. $t$, $e$ and $h$ denote input tokens, encoded EDU representations, and decoded hidden states. The \textit{Stack} is maintained by the decoder to track top-down depth-first span splitting. With each splitting pointer $k$, sub-spans $e_{i:k}$ and $e_{k+1:j}$ are fed to a classifier $\Phi$ for nuclearity and relation determination.}
    \label{fig:Architecture}
\end{figure}

\noindent\textbf{Span Splitting Decoder:}
The decoder splits spans of EDUs to form the tree structure in a top-down transition-based procedure. 
Figure \ref{fig:Architecture} illustrates the parsing steps of the example in Figure \ref{fig:RST_TREE}: the decoder maintains a \textit{Stack}, which is initialized by the span of all EDUs $e_{1:m}$. At each decoding step $t$, the span $e_{i:j}$ at the head of stack is parsed into two sub-spans $e_{i:k}$ and $e_{k+1:j}$ ($i\leq k < j$), and $k$ is the splitting position predicted via an attention-based pointer network \cite{bahdanau2014neural,vinyals2015pointer}. Afterwards, spans containing more than one EDU are pushed into the stack, then the decoder iteratively parses the spans until \textit{Stack} is empty.

\noindent\textbf{Nuclearity-Relation Classifier:}
At each decoding step, after the span $e_{i:j}$ is split into two sub-spans $e_{i:k}$ and $e_{k+1:j}$, a bi-affine classifier \cite{dozat2016deep} is adopted to predict their nuclearity and relation labels. Here we use the joint labels of nuclearity and relation as previous studies \cite{yu2018transition,lin2019unified}.
The total loss is specified as the sum of the cross entropy of span splitting and nuclearity-relation classification. Model implementation details and hyper-parameter configuration are described in Appendix \ref{apdx:hyper-parameter}.

\subsection{Multilingual Parsing}
In this section, we introduce two approaches for the multilingual discourse parsing. Since both methods are model-independent, they can be adopted on various neural architectures, and extended to other low-resource scenarios.

\subsubsection{Utilizing Cross-Lingual Vector Representations}
Recently, the large-scale multilingual language models are able to provide universal encoders that project various inputs to a shared embedding space \cite{Lample-2019-XLM}, and are proved effective in natural language processing tasks such as machine comprehension. Therefore, to conduct discourse parsing on documents from various languages, we first propose to apply a cross-lingual representation backbone in the embedding layer in Section \ref{ssec:neural_parser}. Here, we utilize XLM-RoBERTa \cite{conneau-2020-XLM-Roborta}, which supports 100 languages, and fine-tune it by joint training with the whole neural parser. Moreover, since BERT-based backbones usually have positional embedding limitation, to encode lengthy sequences without truncation for document-level discourse parsing, the sliding window strategy\footnote{In our experiment, the input sequence length after sub-word tokenization can be larger than 2048. In order to exploit global contextual information, the window size is set as 500, and the stride size is 200.} is adopted for better long dependency modeling.

\subsubsection{Adopting EDU Segment-Level Translation}
Aside from using cross-lingual embedding, one alternative way is to transform multilingual text content into a monolingual space. While sophisticated neural approaches are able to generate multilingual translation with high quality and fluency, the commonly adopted sentence-level translation usually makes changes to the syntactic structure, which affects the original discourse annotation like the number and order of EDUs. Therefore, we propose to convert multilingual source content via EDU segment-level translation, as EDU segments are deemed to be semantically cohesive.  We feed the documents with EDU segmentation (split by newlines) to a machine translator \cite{wu2016googleTranslate}, then use the monolingual samples for training and evaluation. As shown in Figure \ref{fig:RST_TREE}, we observe that the translated material retains the original split and order at the EDU level, and shares the same English syntactic characteristics such as the position of discourse connective words (e.g., `however', `although') and relation pronouns (e.g., `that', `which'). Then, we train the neural parser in Section \ref{ssec:neural_parser} on the translated samples with their original tree structure, nuclearity, and relation annotations.

\section{Experimental Results and Analyses}
In this section, we describe data collection and present the experimental setting, results and analyses of the proposed methods.

\subsection{Data and Pre-processing}
\label{ssec:data-processing}
We constructed a multilingual dataset by collecting treebanks from 6 languages: English (En-DT), Brazilian Portuguese (Pt-DT), Spanish (Es-DT), German (De-DT), Dutch (Nl-DT), and Basque (Eu-DT), and their details are shown in Table \ref{data-table}.
Since the annotated formats are slightly different among treebanks, we conducted data pre-processing as in \cite{braud-etal-2017-cross-lingual} to uniform them. All samples were transformed into binary trees, and units that were not linked to others within the tree were removed. Following \cite{lin2019unified}, we reorganized the discourse relations to 18 categories, and attached the nuclearity labels \textit{Nucleus-Satellite (NS)}, \textit{Satellite-Nucleus (SN)} and \textit{Nucleus-Nucleus (NN)} to the relation labels. For each language, we randomly selected 38 samples for evaluation. The total training set and test set are 1.1k and 228. For encoding input, we applied the pre-trained sub-word tokenizer of XLM-RoBERTa \cite{conneau-2020-XLM-Roborta}. We adjusted random seeds to obtain multiple results for each language and used the average as reported scores.

\begin{table}[t]
\linespread{1.2}
\scriptsize
\centering
\begin{tabular}{llccl}
\hline
\textbf{Language}   & \textbf{Tree Bank}                                            & \textbf{Sample Num.} & \textbf{Relation Num.} & \textbf{Material Domain}                                                                                                                              \\
\hline
English (En-DT)    & English RST-DT \cite{carlson2002rst} & 385         & 56            & Wall Street Journal articles                                                                                                        \\
\hline
           & CST-News \cite{cardoso2011cstnews}                    & 140         & 31            & News articles                                                                                                                       \\
Brazilian  & Summ-it \cite{collovini2007-summ-it}                     & 50          & 29            & Science articles                                                                                                                    \\
Portuguese (Pt-DT) & Rhetalho \cite{pardo2005rhetalho}                      & 40          & 23            & Articles in computer science                                                                                                        \\
           & CorpusTCC \cite{pardo2004-CorpusTCC}                  & 100         & 31            & Articles in computer science and news                                                                                          \\
           \hline
Spanish (Es-DT)   & Spanish RST-DT \cite{da2011-spanish-rst-dt}               & 267         & 29            & \begin{tabular}[c]{@{}l@{}}  Text written by specialists on different topics\\ (e.g. astrophysics, economy, law, linguistics)\end{tabular} \\
\hline
German  (De-DT)   & MAZ Corpus \cite{stede2014-MAZ-corpus}  & 175         & 30            & German newspaper commentaries                                                                                                       \\
\hline
Dutch   (Nl-DT)   & Dutch RST-DT \cite{redeker2012-dutch-rst-dt}                                     & 80          & 31            &                     Encyclopedias, letters, and news                                                                                                                \\
\hline
Basque  (Eu-DT)   & Basque RST-DT  \cite{iruskieta2013-basque-rst-dt}             & 88          & 31            & \begin{tabular}[c]{@{}l@{}}Abstracts from three specialized domains\\ (medicine, terminology and science)
\end{tabular}  \\
\hline
\end{tabular}
\caption{The collected RST discourse treebanks from 6 languages.}
\label{data-table}
\vspace{0.1cm}
\end{table}

\begin{table}[t!]
\linespread{1.2}
\scriptsize
\begin{tabular}{l|p{0.34cm}p{0.34cm}p{0.34cm}|p{0.34cm}p{0.34cm}p{0.34cm}|p{0.34cm}p{0.34cm}p{0.34cm}|p{0.34cm}p{0.34cm}p{0.34cm}|p{0.34cm}p{0.34cm}p{0.34cm}|p{0.34cm}p{0.34cm}p{0.34cm}}

\hline
             & \multicolumn{3}{c}{En-DT}                                                 & \multicolumn{3}{c}{Pt-DT}                                                 & \multicolumn{3}{c}{Es-DT}                                                 & \multicolumn{3}{c}{De-DT}                                                 & \multicolumn{3}{c}{Nl-DT}                                                 & \multicolumn{3}{c}{Eu-DT}                                                 \\
             Models         & Sp      & Nu     & Rel    & Sp     & Nu     & Rel     & Sp      & Nu     & Rel    & Sp      & Nu     & Rel    & Sp      & Nu     & Rel    & Sp      & Nu     & Rel    \\
\hline
\hline
\multicolumn{12}{l}{\textbf{MACRO F1 SCORE}} \\
\hline
Human*           & 88.7    & 77.7   & 65.8   & -       & 78.0   & 66.0   & 86.0    & 82.5   & 76.8   & -       & -      & -      & 83.0    & 77.0   & 70.0   & 81.7    & -      & 61.5   \\
MFS*       & 58.2    & 33.4   & 22.1   & 57.3    & 33.9   & 23.2   & 82.0    & 51.5   & 17.7   & 61.3    & 37.8   & 13.2   & 57.9    & 35.5   & 22.0   & 63.2    & 34.9   & 18.8   \\
Li et al. \shortcite{li-etal-2014-recursive}  & 85.0    & 70.8   & 58.6   & -       & -      & -      & -       & -      & -      & -       & -      & -      & -       & -      & -      & -       & -      & -      \\
Braud et al.\shortcite{braud-etal-2017-cross-lingual}    & 85.1    & 73.1   & 61.4   & 81.9    & 65.1   & 49.8   & 88.8    & 68.0   & 50.4   & 79.6    & 53.6   & 34.1   & 69.2    & 43.4   & 28.3   & 76.7    & 50.5   & 29.5   \\
\hline
\multicolumn{3}{l}{Cross-Lingual Representation} \\
EN-Training & 88.1  &  77.3  &  64.7  &  84.9  &  68.6  &  54.7  &  85.2  &  58.1  &  36.8  &  82.0  &  53.5  &  34.5  &  82.3  &  57.7  &  39.4  &  82.4  &  56.5  &  34.7  \\

Multi-Training   & \textbf{88.9} & \textbf{77.5} & \textbf{65.7} & \textbf{87.2} & \textbf{72.9} & \textbf{60.8} & \textbf{88.3} & \textbf{75.3} & \textbf{60.5} & \textbf{84.1} & \textbf{62.8} & \textbf{45.9} & \textbf{86.4} & \textbf{64.6} & \textbf{49.5} & \textbf{85.9} & \textbf{67.3} & \textbf{51.9} \\
\hline
\multicolumn{3}{l}{Segment-Level Translation} \\
EN-Training & 88.3  &  77.8  &  64.9  &  85.1  &  69.2  &  55.0  &  85.9  &  58.6  &  37.0  &  80.7  &  53.3  &  34.1  &  82.6  &  58.0  &  39.7  &  82.6  &  58.3  &  36.2  \\

Multi-Training   & \textbf{89.2} & \textbf{78.7} & \textbf{67.1} & \textbf{87.9} & \textbf{73.8} & \textbf{61.7} & \textbf{89.4} & \textbf{75.8} & \textbf{61.2} & \textbf{82.7} & \textbf{59.5} & \textbf{42.4} & \textbf{85.0} & \textbf{63.4} & \textbf{48.2} & \textbf{85.7} & \textbf{67.8} & \textbf{49.6} \\
\hline
\hline

\multicolumn{12}{l}{\textbf{MICRO F1 SCORE}} \\

\hline
Yu et al. \shortcite{yu2018transition} & 85.5    & 73.1   & 60.2   & -       & -      & -      & -       & -      & -      & -       & -      & -      & -       & -      & -      & -       & -      & -      \\
Iruskieta \shortcite{iruskieta-braud-2019-eusdisparser} & 80.9 & 65.5   & 52.1   & 79.7      & 62.8      & 47.8      & 85.4       & 65.0      & 45.8      & -       & -      & -      & -       & -      & -      & -       & -      & -      \\
\hline
\multicolumn{3}{l}{Cross-Lingual Representation} \\
EN-Training  & 87.2    & 73.7   & 62.3   & 84.4    & 68.1   & 53.9   & 79.5    & 55.6   & 36.0   & 81.7    & 53.1   & 33.8   & 80.5    & 55.6   & 38.5   & 81.7    & 55.3   & 33.6   \\
Multi-Training   & \textbf{87.5} & \textbf{74.7} & \textbf{63.0} & \textbf{86.3} & \textbf{71.7} & \textbf{60.0} & \textbf{86.2} & \textbf{71.1} & \textbf{54.4} & \textbf{83.6} & \textbf{62.2} & \textbf{45.1} & \textbf{85.9} & \textbf{64.5} & \textbf{49.4} & \textbf{85.1} & \textbf{65.8} & \textbf{47.7} \\
\hline
\multicolumn{3}{l}{Segment-Level Translation} \\
EN-Training  & 87.4    & 74.6   & 62.8   & 84.9    & 68.0   & 54.2   & 82.6    & 56.3   & 35.1   & 79.5    & 52.7   & 34.0   & 81.5    & 57.0   & 39.1   & 81.2    & 57.3   & 35.5   \\
Multi-Training   & \textbf{87.8} & \textbf{75.4} & \textbf{63.5} & \textbf{86.5} & \textbf{72.0} & \textbf{60.3} & \textbf{87.9} & \textbf{71.4} & \textbf{56.1} & \textbf{82.3} & \textbf{58.9} & \textbf{41.0} & \textbf{84.6} & \textbf{62.7} & \textbf{47.2} & \textbf{84.4} & \textbf{65.5} & \textbf{47.3} \\
\hline

\end{tabular}
\caption{Evaluation scores on multilingual RST treebanks. * denotes results from \cite{braud-etal-2017-cross-lingual}. \textit{Sp}, \textit{Nu} and \textit{Rel} denote span splitting, nuclearity and relation determination respectively.}
\label{result-table}
\end{table}

\subsection{Evaluation Result}
The experimental results are shown in Table \ref{result-table}. Since macro-averaged and micro-averaged F1 scores are reported in different previous works, we conducted extensive comparisons using these two criteria. The results demonstrate that (1) models which are only trained on the English treebank (\textit{EN-Training}) can achieve competitive span splitting performance on the multilingual test sets; 
(2) the two proposed approaches with multilingual training (\textit{Multi-Training}) surpass the baselines with a significant margin at all fronts: the span splitting prediction on all languages are approaching human performance, and nuclearity and relation determination are improved substantially compared to previously reported cross-lingual parsers \cite{braud-etal-2017-cross-lingual}; 
(3) Interestingly, the model with cross-lingual representation performs slightly better on the treebanks with fewer samples (e.g., De-DT, Nl-DT, and Eu-DT), and the model with segment-level translation obtains the best result in English.

\subsection{Topic Modeling Analysis}
\begin{figure}[t]
    \centering
    \includegraphics[width=16cm]{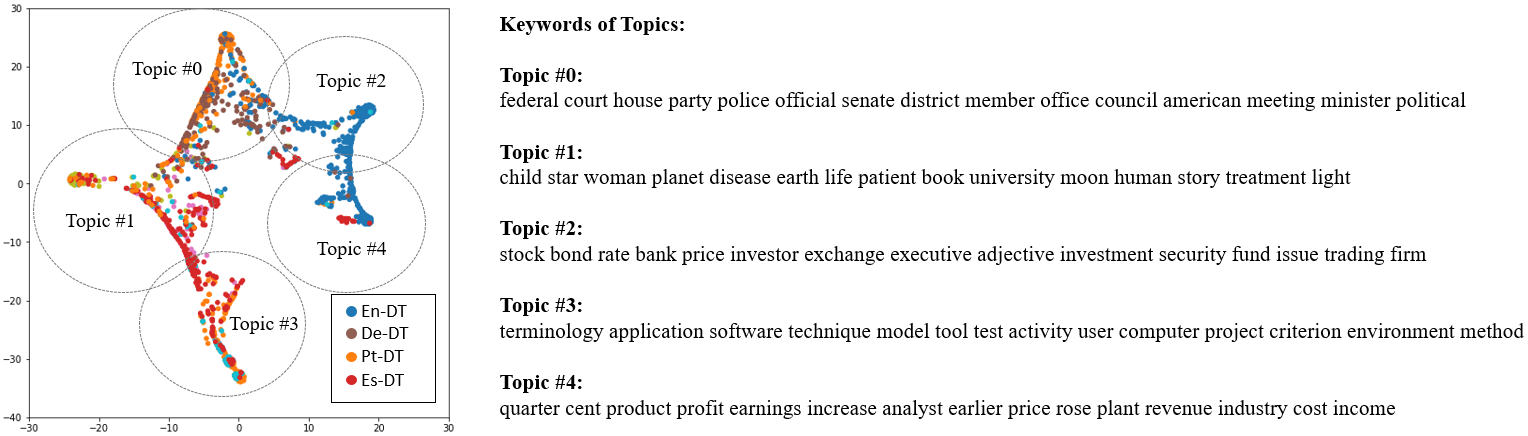}
    \caption{Analysis with topic modeling and feature visualization. Topic keywords are extracted with the LDA algorithm, and the scatter plot is illustrated by applying t-SNE.}
    \label{fig:topic_analysis}
\end{figure}

To further assess the generality of our parsers from the domain perspective, we conducted a topic modeling analysis on the translated samples from multilingual treebanks. LDA (the topic number was set as 5) and t-SNE were used for topic modeling and feature visualization, respectively. As shown in Figure \ref{fig:topic_analysis}, the treebanks show a trend to cluster in different topics (marked in circles). For instance, the English treebank (En-DT) mainly focuses on the financial news domain (in blue). Compared to the Portuguese treebank (Pt-DT), the Spanish one (Es-DT) is more distinct to the En-DT, which is consistent with the performance gap between them under \textit{EN-Training} (see Table \ref{result-table}). Therefore, by adding Spanish (Es-DT) and Portuguese (Pt-DT) data, topic coverage for the \textit{Multi-Training} model expands to scientific and terminology articles, and thus becomes more generalizable to other domains.

\section{Conclusion}
In this paper, we investigated two approaches for cross-lingual neural discourse parsing. Experimental results show that both utilizing cross-lingual representation and adopting segment-level translation contribute to obtaining state-of-the-art performance on various treebanks. Moreover, monolingual models can also benefit from cross-lingual training by introducing data from more domains. For future work, we consider conducting domain adaption via few-shot learning to make our approach more generalizable.

\section*{Acknowledgements}
This research was supported by funding from the Institute for Infocomm Research (I2R) under A*STAR ARES, Singapore. We thank N. Asher, A. T. Aw, S. Joty, W. Lei, B. Webber for insightful discussions, C. Braud for sharing linguistic resources, and the anonymous reviewers for their precious feedback to help improve and extend this piece of work.



\bibliographystyle{coling}
\bibliography{coling2020}

\clearpage

\appendix

\section{Model Description and Hyper-parameter Configuration}
\label{apdx:hyper-parameter}

\subsection{Details of Encoder-Decoder}
Given a document containing $n$ tokens $T=\{t_1, t_2,...,t_n\}$, the embedding layer (a pre-trained language model) produces token-level embeddings $\widetilde{T}=\{\widetilde{t}_1,...,\widetilde{t}_n\}$, the EDU-level representations $C = \{c_1,...,c_m\}$ are calculated by averaging the respective token-level embeddings. Then, a multi-layer Bi-GRU is employed to generate the context-aware EDU-level representations $V=\{v_1,..., v_m\}$ by sequentially modeling the dependency among $C$, and each $v_i$ is the concatenated vector of the the forward and the backward hidden states: $v_i=[\stackrel{\rightarrow}{v_i}; \stackrel{\leftarrow}{v_i}]$. Afterwards, the final EDU representations are produced via incorporating boundary embeddings at the beginning and end of each EDU from $\widetilde{T}$ to the context-aware EDU vector $v_i$:
\begin{equation}
\label{boundary_equation}
    e_i = W_e([v_i;\widetilde{t}_{i\_start};\widetilde{t}_{i\_end}]) + b_e
\end{equation}
where $;$ denotes the concatenation operation. $W_e$ and $b_e$ are the trainable parameter matrix and bias.

We employ a unidirectional GRU layer for the span splitting decoder, and its hidden state $h_0$ is initialized by the last hidden states of the encoder.  At each decoding step, the hidden state $h_t$ is produced by the GRU with the previous hidden state $h_{t-1}$ and the input span representation $e_{i:k}$, where $e_{i:k}$ is calculated from taking the average of the respective EDU representations (\emph{i.e.} $mean(e_i,...,e_k)$ for $e_{i:k}$). Then, the pointer network \cite{vinyals2015pointer} is used to predict the splitting position according to the computed attention scores on encoded EDU representations, which is a softmax distribution over the input span.
\begin{equation}
    s_{t,u} = \sigma (h_t, e_u) \ \  \mathbf{for}\ \  u = i...j
\end{equation}
\vspace{-0.5cm}
\begin{equation}
    a_t = softmax(s_t) = \frac{exp(s_{t,u})}{\sum_{u=i}^jexp(s_{t,u})}
\end{equation}
\noindent where $\sigma(x,y)$ is the dot product used as attention scoring function.

\subsection{Details of Nuclearity-Relation Classifier}
After decoder splits span $e_{i:j}$ into left sub-span $e_{i:k}$ and right sub-span $e_{k+1:j}$, the classifier first projects $e_l$ and $e_r$ to latent features $\widetilde{e}_l$ and $\widetilde{e}_r$ by a linear layer with Exponential Linear Unit (ELU), where $e_l$ and $e_r$ are the average of respective EDU representations in $e_{i:k}$ and $e_{k+1:j}$:
\begin{equation}
    \widetilde{e}_l=ELU(e_l^TU_1);\  \widetilde{e}_r = ELU(e_r^TU_2)
\end{equation}
Then a bi-affine layer with softmax activation maps the features to nuclearity-relation labels:
\begin{equation}
\begin{aligned}
    P_\theta(y|X) = softmax(\widetilde{e}_l^TW_l + \widetilde{e}_l^TW_{lr}\widetilde{e}_r + \widetilde{e}_r^TW_r + b)
\end{aligned}
\end{equation}
\noindent where $W_l\in\mathcal{R}^{d\times R}$; $W_r\in\mathcal{R}^{d\times R}$ and $W_{lr}\in\mathcal{R}^{d\times d\times R}$ are the weights and bias $b \in \mathcal{R}^R$.

\subsection{Training Loss}
The parser's objective contains two parts: building the discourse tree structure and predicting the nuclearity and discourse relation labels. Therefore, the total loss is the sum of structure loss $\mathcal{L}_s$ and label prediction loss $\mathcal{L}_l$, where $\mathcal{L}_s$ is the cross entropy loss upon attention probabilities of the pointer network and $\mathcal{L}_l$ is the cross entropy loss of the nuclearity-relation classification. 

\begin{equation}
    \mathcal{L}_s(\theta_s) = - \sum_{t=1}^TlogP_{\theta_s}(y_t|y_1,...,y_{t-1},X)
\end{equation}
\vspace{-0.2cm}
\begin{equation}
    \mathcal{L}_l(\theta_l) = - \sum_{m=1}^M\sum_{r=1}^RlogP_{\theta_l}(y_m=r|X)
\end{equation}
\noindent where $\theta_s$ and $\theta_l$ are the parameters of the pointer network and classifier respectively, $T$ is the total number of spans, and $y_1,...,y_{t-1}$ denote the sub-trees that have been generated in the previous steps. $M$ is the number of spans that need to be split, and $R$ is the number of nuclearity-relation labels. \\
The total loss with $L_2$-regularization is:
\begin{equation}
     \mathcal{L}_{total}(\theta^{*}) = \mathcal{L}_s(\theta_s) + \mathcal{L}_l(\theta_l) + \lambda||\theta^{*}||_2^2
\end{equation}
\noindent where $\lambda$ is the regularization strength and $\theta^{*}$ refers to all learning parameters of the model.

\subsection{Hyper-parameter Configuration}
The neural model was implemented in PyTorch \cite{paszke2019pytorch}. We used `xlm-roberta-base' implemented in \cite{Wolf2019HuggingFacesTS} and fine-tuned the last 4 layers during training. In order to exploit global contextual information, the window size was set as 500 and the stride size was 200. Documents were tokenized via the sub-word scheme as in \cite{Lample-2019-XLM}. We trained the model for 30 epochs and selected the best checkpoints on a validation set for evaluation. Adam optimization algorithm was used with batch size of 3, weight decay of 5e-5, and learning rate of 1e-4. Dropout rate was set as 0.5 during training. The embedding dimension and hidden size were 768 and 384. The trainable parameter size was 67M, where 31M parameters were from fine-tuning the language model. All experiments were run on a Tesla V100 GPU with 16GB memory. 

\end{document}